\documentclass[lettersize,journal]{IEEEtran}
\usepackage{amsmath,amsfonts}
\usepackage{algorithmic}
\usepackage{algorithm}
\usepackage{array}
\usepackage[caption=false,font=normalsize,labelfont=sf,textfont=sf]{subfig}
\usepackage{textcomp}
\usepackage{stfloats}
\usepackage{url}
\usepackage{verbatim}
\usepackage{graphicx}
\usepackage{cite}

\begin{document}

\title{Denoising of Geodetic Time Series Using\\Spatiotemporal Graph Neural Networks: Application to Slow Slip Event Extraction}

\author{Giuseppe Costantino,
        Sophie Giffard-Roisin,
        Mauro Dalla Mura,
        and Anne Socquet

\thanks{Giuseppe Costantino, Sophie Giffard-Roisin and Anne Socquet are with Univ. Grenoble Alpes, CNRS, IRD, ISTerre, 38000 Grenoble, France.\\
Mauro Dalla Mura is with Univ. Grenoble Alpes, CNRS, Grenoble INP, GIPSA-lab, 38000 Grenoble, France and Institut Universitaire de France, Paris, France.}
}

\maketitle

\begin{abstract}


Geospatial data has been transformative for the monitoring of the Earth, yet, as in the case of (geo)physical monitoring, the measurements can have variable spatial and temporal sampling and may be associated with a significant level of perturbations degrading the signal quality.
Denoising geospatial data is, therefore,
essential, yet
often challenging because the observations may
comprise noise coming from different origins, including both environmental signals and instrumental artifacts, which are spatially and temporally correlated, thus hard to disentangle. This study addresses the denoising of multivariate time series acquired by irregularly distributed networks of sensors,
requiring specific methods to handle
the spatiotemporal correlation of the noise and the signal of interest. Specifically, our method focuses on the denoising of geodetic position time series,
used to monitor ground displacement worldwide with centimeter-to-millimeter precision. Among the signals affecting
GNSS data, slow slip events (SSEs) are of interest to seismologists. These
are transients of deformation that are weakly emerging
compared to other signals. Here, we design SSEdenoiser, a multi-station spatiotemporal graph-based attentive denoiser that learns latent characteristics of GNSS noise to reveal SSE-related displacement with sub-millimeter precision. It is based on the key combination of graph recurrent
networks and spatiotemporal Transformers. The proposed method is applied to the Cascadia subduction zone, where SSEs
occur along with bursts of tectonic tremors, a seismic rumbling identified from independent seismic recordings. The extracted
events match the
spatiotemporal evolution of tremors.
This good space-time correlation of the denoised GNSS signals with the tremors
validates the proposed denoising procedure.

\end{abstract}


\begin{IEEEkeywords}
Time series analysis, denoising, deep learning, graph neural networks, multi-station, spatiotemporal, spatiotemporal attention, geospatial data, GNSS (Global Navigation Satellite System), GPS (Global Positioning System), slow slip events, SSEs (slow slip events), geodesy, seismology
\end{IEEEkeywords}

\section{Introduction}

    \IEEEPARstart{G}{eospatial} data represents a crucial resource for the efficient acquisition of large-scale measurements, providing a detailed view of Earth processes across numerous applications. In the past few decades, the use of geospatial data (\textit{e.g.}, satellite imagery, Global Navigation Satellite System (GNSS), seismic data, climate data) has significantly contributed to remote sensing, both for the monitoring of the environment \cite{luo2023multiscale} and natural processes \cite{hong2007experimental, yi2020new, wagner2012geospatial} and for urban \cite{ding2022bi} and infrastructure development \cite{benediktsson2005classification, ghamisi2014fusion}. Geospatial data typically incorporates information on the location of target objects with the temporal tracking of some of their characteristics, allowing for spatiotemporal analysis. However, these measurements are strongly affected by various perturbations. These noises can be spatially and temporally correlated, making it challenging to extract useful information, \textit{i.e.}, to separate the characteristics of target objects from those of external perturbations. Depending on the specific application, the signal can be modeled as a mixture of several sources, \textit{e.g.}, atmospheric interference, uncertainties and artifacts related to the sensor measurements, and different tectonic or topographic signals. Separating these mixed sources can be a difficult task, yet it is key to performing successful signal extraction. Depending on the structural type of geospatial data (\textit{e.g.}, images, time series), different noise components arise, which require dedicated post-processing, such as in the case of image-like data, \textit{e.g.}, satellite images, or time-series-like data, \textit{e.g.}, Global Navigation Satellite System (GNSS) data.  In this paper, we focus on time-series-like data and, more specifically, we address the problem of denoising multivariate time series in a multi-station setting. We target our analysis on the denoising of GNSS position time series and we apply our method to the extraction of transients of aseismic deformation. Albeit designed for this specific task, the developed framework is general enough to be applied to any time series denoising problem with multiple measurement sites (array of sensors) and multiple non-independent components or channels (multivariate), \textit{e.g.}, GNSS data, seismic data, gravimeters, tiltmeters, strainmeters.

    GNSS is an essential tool in modern geodesy, providing accurate and precise positional information that has transformed our understanding of Earth's shape, movement, and processes. Using satellite constellations, time series of displacement relative to a GNSS site (antenna) are calculated, which can have high precision, usually of the order of the millimeter. GNSS time series usually fall into two categories: high-rate GNSS, typically sampled at 5-minute intervals, and daily GNSS, where measurements are averaged to provide a daily displacement value. GNSS sites are arranged as a network of antennas, resulting in uneven regional sampling that is sparser than satellite images, for instance.
    
    The use of the GNSS has been transformative for various remote sensing applications, such as the observation and monitoring of soil moisture \cite{camps2016sensitivity, pierdicca2021potential}, landslide monitoring \cite{dai2020entering} and hydrology-related applications \cite{unwin2021introduction}, opening a large number of perspectives both towards societal and environmental advancements \cite{garrison2014ieee, zavorotny2014tutorial}. Also, several studies focus on improving the precision of the GNSS positioning \cite{li2014high} as well as extracting statistics from the noise affecting the data \cite{montillet2012extracting}. Recently, advanced signal processing methods (among which machine learning) have been explored for subsurface exploration with promising advances towards better imaging the Earth structure \cite{alregib2018subsurface}.
    In this direction, deep-learning-based methods have been widely used in remote sensing as a way to boost traditionally employed strategies as well as to extract finer-grained knowledge from massive data sets. Diverse studies have used machine and in particular deep learning in Geoscience, \textit{e.g.}, for the detection of volcano-seismic signals \cite{malfante2018machine, peixoto2021tensor, lara2020automatic, lopez2020contribution}, seismic signal restoration \cite{chai2020deep} and full-waveform inversion \cite{zhang2021deep}. Some works also use GNSS data, \textit{e.g.}, for probabilistic source characterization \cite{lin2023hybrid} or time series regression \cite{shahvandi2021modified}. However, the number of studies using GNSS data for tectonics is still limited, probably due to its scarcity (in space, but also in time: the acquisition frequency is between an hour and one day) and because proper modeling of the noise in the GNSS positioning remains a challenge.
    
    Position time series derived from GNSS are affected by different signals generally characterized by spatial and temporal correlation. The challenge in GNSS data denoising lies in developing a method able to learn how to decorrelate these signals by separating what we consider as noise and the different signals from each other. Two major families of signals can be identified: (1) geophysical (\textit{e.g.}, tidal or hydrological loading) and tectonic signals (\textit{e.g.}, earthquakes and related processes, such as aseismic deformation and post-seismic relaxation) and (2) uncertainties in the position (\textit{e.g.}, errors on the calculation of the orbits of the satellites, clock synchronization, multipath errors, ionospheric and tropospheric delays \cite{enge1994global, wdowinski1997southern, williams2004error}). Most studies use parametric models at each station \cite{marill2021fourteen, bedford2018greedy} or blind source separation methods \cite{michel2019interseismic} to model signals constituting the GNSS position, yet these methods may not be able to fully reproduce the complexity of the signals.

    Recent works have shown promising results in seismology to address the problem of denoising seismic waveforms or high-rate GNSS time series for seismological applications. Saad et al. \cite{saad2020deep} developed a deep autoencoder to attenuate the random noise affecting time-series seismic data. The model works by first encoding the time series to several levels of abstraction and then decoding the compressed information to reconstruct the noiseless seismic signal. Zhou et al. \cite{zhu2019seismic} designed a deep-learning-based decomposition/denoising method named DeepDenoiser. The method is a U-Net trained to learn a sparse representation of the data in the time-frequency domain and nonlinear-mapped masks to separate the signal of interest from the noise. Thomas et al. \cite{thomas2023deep} use a modified version of DeepDenoiser adapted to work with 3-component high-rate GNSS data. They test three versions of DeepDenoiser to account for amplitude distortions and phase differences. All the previous methods employ single-station time series as input, to provide clean time series for each station separately. However, the main limitation arising from the use of single-station methods is that the spatial variability of the data is not exploited. Multi-station methodologies have been also explored in seismology as an attempt to take into account the spatial coherency of the measurements, for several tasks in which this would translate into an improvement of the performance \cite{munchmeyer2021earthquake, van2020automated}.
    
    In the case of GNSS data, and especially for the identification of slow slip events, exploiting the spatial variability of the GNSS position becomes critical because of the lower signal-to-noise ratio compared to seismic data or high-rate GNSS. Moreover, both the GNSS noise perturbations and the interesting tectonic signals are spatially and temporally correlated and, therefore, the spatial and temporal analysis should not be decoupled. In previous works, we tested several GNSS data representations aiming to explicitly inject spatial information in the context of earthquake characterization \cite{costantino2023seismic} and we developed a multi-station attentive deep neural network for slow slip event identification \cite{costantino2023multi}. We found that multi-station approaches achieve better performance when targeting the ground deformation with daily position GNSS time series and should be generally preferred over single-station methods.

    The aforementioned approaches rely on single-station analysis, which represents a limitation for GNSS, being the measured displacement correlated in the spatial and temporal dimensions. In this work, we aim to develop a deep-learning-based method for the denoising of raw GNSS data with a multi-station approach. One way to deal with sparse GNSS measurements is to arrange them as a matrix, where each row is a time series, sorted for example by latitude (or longitude), and to use 2D Convolutional Neural Networks (CNNs), specifically developed for image-like data (see the representation in Figure \ref{fig:data}(a)). With this method, two-dimensional convolutions (time and station dimension) can model both local- and large-scale spatial relationships between stations, but no explicit information coming from the geometry of the GNSS network is enforced: one out of the two spatial dimensions is lost. To this end, spatiotemporal Graph Neural Network (STGNN) methods \cite{yu2017spatio}, which, to the best of our knowledge, have never been applied to remote sensing, can learn these relationships and take full benefit of the spatial information, with the potential of outperforming the 2D-CNN approach. However, developing such models requires setting up explicit joint processing of the information in the temporal and spatial domains. Classical graph-based approaches rely on message passing to propagate the information between neighbouring nodes. Time-varying features can be handled by using a Recurrent Neural Network (RNN) cell as an aggregation function, \textit{e.g.}, long short-term memory (LSTM) or gated recurrent unit (GRU), but extending this mechanism to multivariate time series analysis is not straightforward since GNSS measurements usually have more than one temporal channels (here, we use North-South displacement and East-West displacement components). One solution is to consider the different components as statistically independent and enforce classic message-passing approaches on each component independently. Yet, the assumption of independence between components is too strong since the different GNSS components are generally correlated.
    
    Spatiotemporal approaches have thus been developed to jointly rely on multiple sensor measurements and multiple components. They can be classified in RNN-based and attention-based methods, and are capable of dealing with 3D data (here, [stations, time, directions]). The first family of approaches uses RNNs to extract temporal features \cite{bai2020adaptive}, while the second focuses on attention mechanisms in the time and/or space dimension \cite{shi2019two}. Also, spatiotemporal methods usually consider that the graph connectivity (adjacency matrix) is available beforehand. Most of the previous works compute the adjacency matrix based on distance metrics (\textit{e.g.}, traffic forecasting) \cite{yu2017spatio} or prior information on the nodes (\textit{e.g.}, skeleton-based action recognition) \cite{shi2019two}. However, relying on pre-computed connectivity might not be the optimal choice, since additional edges could be learned from the data, and superfluous connections could also be removed. Moreover, graph connectivity is not always available, as in our case, and computing a satisfactory adjacency matrix is challenging since it would require connecting nodes both within a homogeneous mesh and with long-range connections, which are not easy to model.
    Hence, the solution of learning the adjacency matrix during the training, following the same approach as Bai et al. \cite{bai2020adaptive}, is often relevant.

    In this paper, we apply the denoising of GNSS position time series to the identification of slow slip events. These tectonic events are characterized by slip on faults, as are earthquakes, yet, unlike them, they do not produce ground shaking and they last over days to years. However, most of the occurrences of slow slip transients have low amplitude and remain undetected, since other signals (either geophysical signals or noise) prevail in the GNSS time series. Thus, this represents a relevant case study for the denoising of GNSS data, which can be extended to many other applications dealing with the extraction of low-amplitude signals mixed with other noise components.

    Here we present SSEdenoiser, a spatiotemporal graph-based deep neural network designed to extract spatiotemporal features from multi-station raw (daily) GNSS position time series and to leverage them to isolate the aseismic slip contribution (SSEs) from the rest of the tectonic and non-tectonic signals.
    The originality and novelty of this work lie in (1) the coupled processing of the spatial and temporal information through a spatiotemporal graph neural network with a learned graph connectivity, aimed to uncover hidden relationships between GNSS sites based on the GNSS recordings themselves and (2) the use of a spatiotemporal Transformer, aimed to filter the relevant information encoded by the graph neural network both in space and time.

    The rest of the paper is organized as follows. In section \ref{sec:methodology}, we present the methodology. We first introduce our overall approach, then we describe the data generation strategy and the architecture of the proposed model. In section \ref{sec:results}, we present the experimental results, in terms of: results on the synthetic data set (with a comparison with several baseline methods), an ablation study to test the robustness of the proposed methodology, analysis of the learned graph connectivity, results on real GNSS data in the Cascadia subduction zone and qualitative comparison between the proposed approach and a previous single-station slow slip event detector for the Cascadia subduction zone.
    
\section{Methodology} \label{sec:methodology}

    \subsection{Overall approach}

        In this study, we build SSEdenoiser, a multi-station deep-learning-based method aimed at denoising raw GNSS position time series to extract aseismic transients of deformation. We train SSEdenoiser on synthetic data because of the paucity of previously catalogued slow slip events, by following the same approach previously designed for slow slip event detection \cite{costantino2023multi}. We generate synthetic data by relying on an underlying additive model. The displacement $\boldsymbol{\xi}_i(t) \in \mathbf{R}^{N_c}$ at a given station $s_i$, with $N_c$ the number of components, is obtained as the sum of a noise term $\mathbf{n}_i(t)$ and a synthetic slow slip signal $\mathbf{d}_i(t)$:

        \begin{equation}
            \boldsymbol{\xi}_i(t) = \mathbf{n}_i(t) + \mathbf{d}_i(t)
        \end{equation}

        We generate the noise term (here, everything that is not the displacement of interest) $\mathbf{n}(t)$ from raw GNSS time series using a Principal Component Analysis (PCA) and a Fourier phase randomization and amplitude matching technique, as described in \cite{costantino2023multi}. As a result, the generated noise will have the same spatial covariance and the same spectral and statistical characteristics as the real noise. We generate the synthetic slow slip signal $\mathbf{d}(t)$ by relying on physical models and by drawing random events along the subduction interface, with various nuances of signal-to-noise ratio and randomly generated physical parameters. We design the database to contain four different settings: (0 events, 1 event, 2 events, 3 events). The final database contains 25\% of each setting. We detail the data generation strategy in the following paragraph.

        SSEdenoiser is a deep denoising neural network, made of the combination of a graph-based recurrent neural network, extracting spatial and temporal features from the data in a multi-station fashion, and a spatiotemporal Transformer, designed as a cascade of temporal and spatial self-attention, capable of focusing on precise space-time relationships. We train the method to extract the slow slip event signals in a temporal window with a fixed size of 60 days. We test the method both on synthetic and real data. On synthetic data, we feed 60-day samples to SSEdenoiser to test the performance both on \textit{negative} samples (0 events in the window) and on \textit{positive} samples (up to three events in the window). On real data, we use a sliding window approach, by collecting the denoised data for each window and by aggregating them afterwards.

    \subsection{Synthetic data generation}

        \begin{figure*}[!t]
            \centering
            \includegraphics[width=\linewidth]{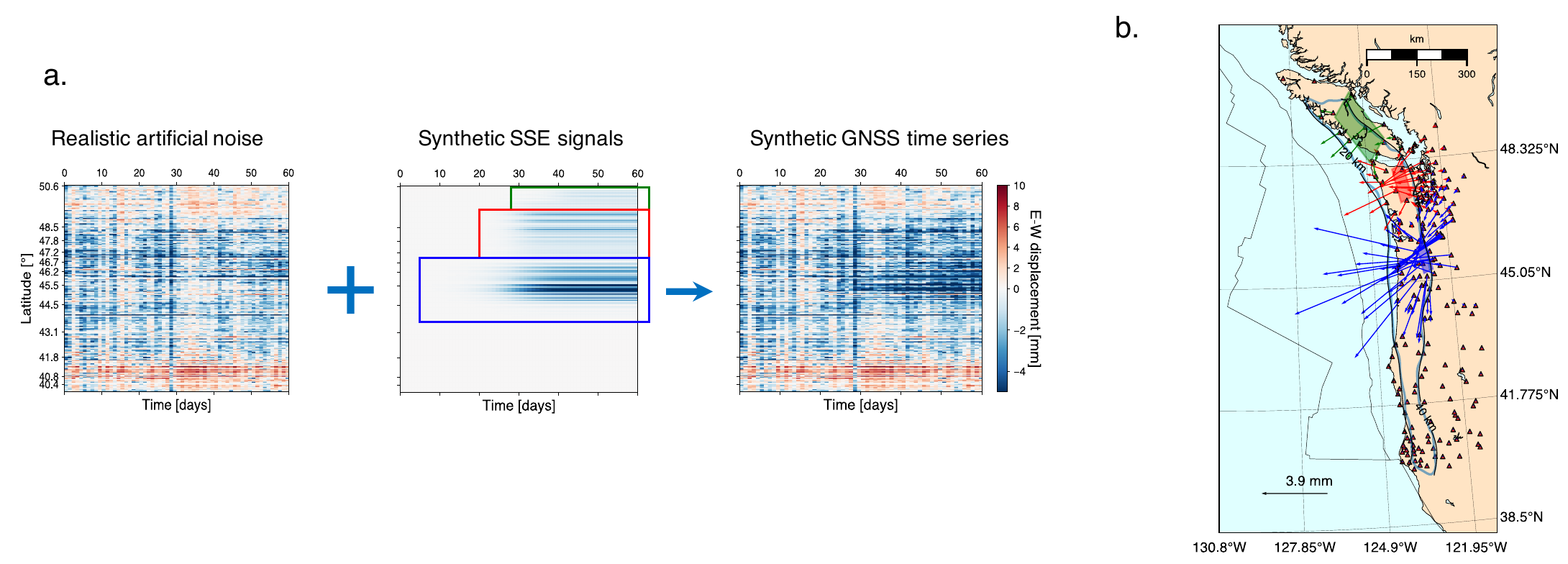}
            \caption{Overview of the synthetic data generation. (a) Each row of the matrices represents a synthetic detrended GNSS position time series (E-W component) color-coded by the amplitude of displacement, created using the proposed synthetic data generation technique. Starting from the left: artificial noise $n(t)$, modeled SSE signals $d(t)$, synthetic GNSS time series $\xi(t)=n(t)+d(t)$. (b) The red triangles represent the location of the GNSS stations (MAGNET GNSS network) used in this study. The arrows show the static synthetic displacement modeled at each GNSS station. In this example, the SSE signal is modeled through three dislocations, shown with different colors (red, green, blue), slipping in an elastic half-space with a different slip amount, slip initiation time, and slip duration. The synthetic displacement time series associated with each dislocation are identified by rectangles in the $d(t)$ matrix in the (a) panel with the same color. The light blue contour represents the SSE area, \textit{e.g.}, the locations of the generated synthetic slow slip events.}
            \label{fig:data}
        \end{figure*}
        
        In \cite{costantino2023multi}, we introduced SSEgenerator, a method for generating synthetic GNSS noise time series and synthetic slow slip events, based on synthetic surface displacements output by physical models. Here we improve and adapt this method for a denoising pipeline and we detail the technique in the following paragraphs. We present an overview of the synthetic data generation in Figure \ref{fig:data}.

        We use 200 raw GNSS position time series from the Cascadia subduction zone from 2007 to 2022. We choose the stations with the least missing data \cite{costantino2023multi}. As opposed to SSEgenerator, we use the full 15-year sequence to generate the geodetic noise, to have longer sequences and thus a more realistic artificial noise time series. Moreover, we allow each sample to belong to one out of four different settings. 25\% of the synthetic samples are considered as \textit{negative} samples, thus they do not contain any SSE signal. These are achieved by training the model to output a zero-displacement signal, thus focusing on extracting an enhanced representation of the input noise. When addressing denoising, this is key to learning the noise structure at best and can be used to better denoise the \textit{positive} samples. 

        The challenge in constructing the slow slip signal $\mathbf{d}_i(t)$ lies in employing a synthetic displacement model that generates surface displacements that mimic well the features expected for slow slip events. To do so, we use Okada \cite{okada1985surface} equations, that link the slip on a dislocation buried in an elastic medium to surface displacements. This allows us to generate synthetic surface displacements that meet the expected physical properties of slow slip events. Because SSEs can have a complex source, including variable slip or lateral propagation of the slip pulse at depth, we impose, for each 60-day window, a variable number of synthetic dislocations per window between 0 and 3 (25\% each), with independent source parameters
        
        
        We uniformly generate the SSE sources within a band along the subduction between 20 and 40 km depth (see Figure \ref{fig:data}(b)), with uniformly generated source parameters (magnitudes uniformly drawn between 6 and 7, strike and dip angles enforced by the slab geometry with $\pm 10$ km variability on the depth and rake angle variable from 80 to 100 degrees to produce a thrust focal mechanism). For more details, we refer the reader to our previous work \cite{costantino2023multi}.
        The temporal signature of the SSE signals is assumed to be a logistic function. We allow the SSEs to last between 10 and 30 days. We model the SSE temporal evolution (one component is shown for simplicity) as follows:

        \begin{equation}
            d_i(t) = \frac{D}{1 + e^{-\beta (t-t_0)}}
        \end{equation}

        where $D$ is the static displacement output by the dislocation model \cite{okada1985surface}, $\beta$ is associated with the growth rate of the curve and $t_0$ is the time corresponding to the inflection point of the logistic function. The value of $D$ modulates the amplitude of the surface displacement and depends on the source parameters (\textit{e.g.}, magnitude, depth). By generating surface displacements corresponding to randomly generated source parameters, we explore different nuances of signal-to-noise ratio to train the denoising method.
        We generate $t_0$ randomly between 0 and 60 days. We obtain $\beta$ as a function of the slow slip event duration $T$, which can be rewritten as $T=t_{max} - t_{min}$, where $t_{max}$ is the time corresponding to the post-SSE displacement (\textit{i.e.}, $D$), and $t_{min}$ is the pre-SSE displacement (\textit{i.e.}, 0). We set a threshold $\gamma$, such that $t_{max}$ and $t_{min}$ are associated with $d_s(D-\gamma D)$ and $d_s(\gamma D)$, respectively, and $\gamma=0.01$. We finally obtain:
        
        \begin{equation}
            \beta = \frac{2}{T} \ln{\left(\frac{1}{\gamma} - 1\right)}.
        \end{equation}

        We generate the slow slip event duration $T$ as a uniform random variable from 10 to 30 days. We further extract 60-day windows of noise $n_i(t)$ to be added to the slow slip signals $d_i(t)$. We also account for missing data by relying on the data gap distribution of the real data.
        

    \subsection{Architecture of SSEdenoiser}

        \begin{figure*}[!t]
            \centering
            \includegraphics[width=\linewidth]{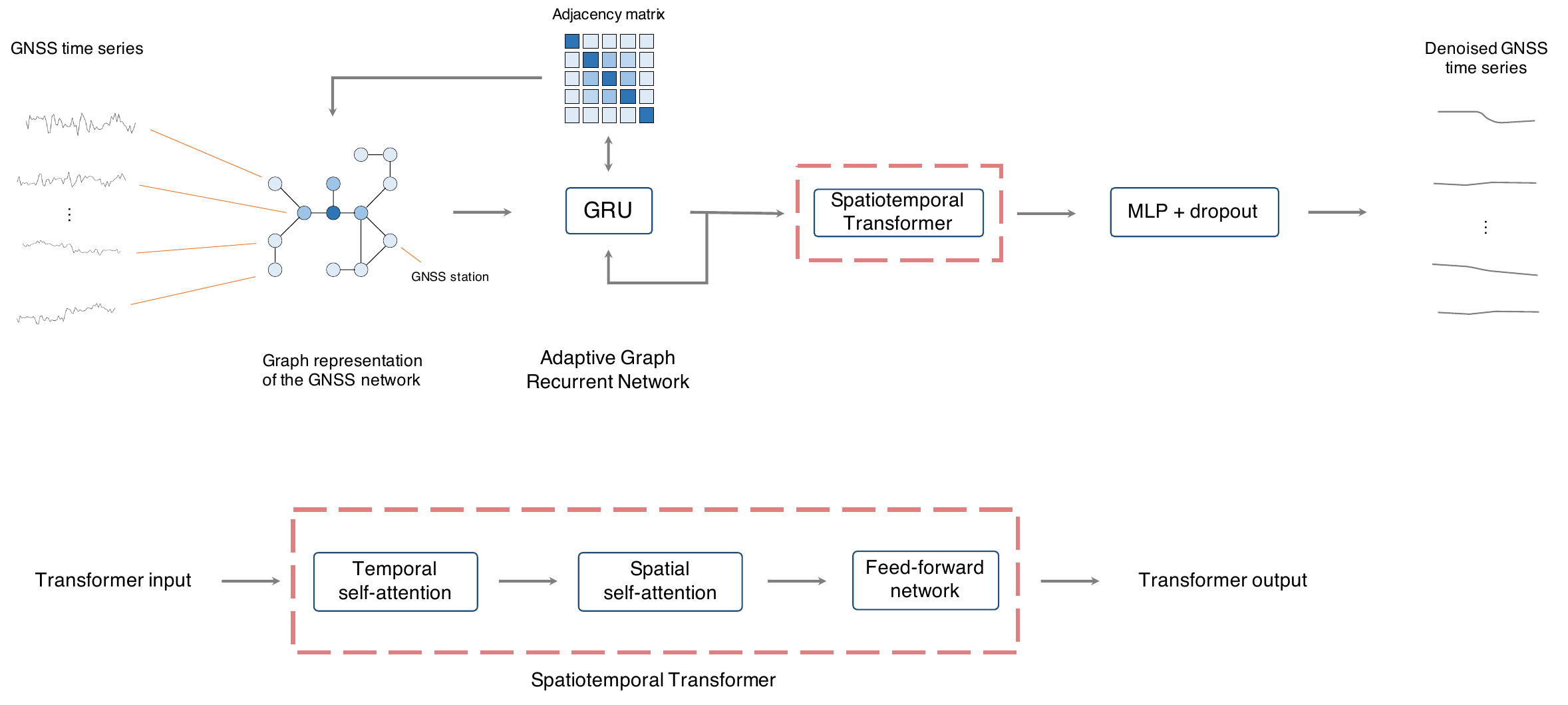}
            \caption{High-level architecture of SSEdenoiser. GNSS time series are first processed by a graph-based recurrent neural network, where temporal features are extracted and spatial relationships are inferred by learning the adjacency matrix. A spatiotemporal Transformer is then used, where temporal and spatial self-attentions attend to the learned temporal features and the spatial relationships. At the end of the pipeline, a fully connected network reprojects the Transformer's output to the input dimension to produce the denoised GNSS time series.}
            \label{fig:sse-denoiser}
        \end{figure*}

        SSEdenoiser is a deep neural network consisting of two main modules: a graph-based recurrent neural network and a spatiotemporal transformer. A high-level architecture is provided in Figure \ref{fig:sse-denoiser}.

        The graph-based recurrent neural network module has been taken from the work by Bai et al. \cite{bai2020adaptive}. The method relies on a learnable adjacency matrix $A$, computed as:

        \begin{equation}
            A = \text{softmax}(\text{ReLu}(\mathbf{E}\mathbf{E}^T))
        \end{equation}

        where $\mathbf{E} \in \mathbb{R}^{N \times d_N}$ are learnable node embeddings. $N$ is the number of nodes in the graph and $d_N$ is the node embedding dimension. We set $N=200$, corresponding to the number of GNSS stations, and $d_N=32$. The adjacency matrix is learned to be symmetric, as in the case of undirected graphs. The adjacency matrix is used by the graph convolutional recurrent unit, having the following constitutive equations \cite{bai2020adaptive}:

        \begin{equation}
            \mathbf{z}(t) = \sigma(A [\mathbf{X}(t), \mathbf{h}(t-1)]\mathbf{E}\mathbf{W_z} + \mathbf{E}\mathbf{b_z})
        \end{equation}

        \begin{equation}
            \mathbf{r}(t) = \sigma(A [\mathbf{X}(t), \mathbf{h}(t-1)]\mathbf{E}\mathbf{W_r} + \mathbf{E}\mathbf{b_r})
        \end{equation}

        \begin{equation}
            \mathbf{\hat h}(t) = \tanh{(A [\mathbf{X}(t), \mathbf{r} \odot \mathbf{h}(t-1)]\mathbf{E}\mathbf{W_{\hat h}} + \mathbf{E}\mathbf{b_{\hat h}})}
        \end{equation}

        \begin{equation}
            \mathbf{h}(t) = \mathbf{z} \odot \mathbf{h}(t-1) + (1 - \mathbf{z}) \odot \mathbf{\hat h}(t)
        \end{equation}
        
        where $\mathbf{X}(t)$ and $\mathbf{h}(t)$ are the input and output at time $t$, $[\cdot]$ is the concatenation operation, $\sigma(\cdot)$ the sigmoid function and $\odot$ the Hadamard (element-wise) product. $\mathbf{W_z}$, $\mathbf{W_r}$, $\mathbf{W_{\hat h}}$, $\mathbf{b_z}$, $\mathbf{b_r}$ and $\mathbf{b_{\hat h}}$ are learnable parameters. We set the hidden size of the recurrent unit to 128.

        The output of the first module, \textit{i.e.}, the output of the last hidden layer $\mathbf{h}(t)$ is then used as input of a Transformer neural network. Usually, transformers are used to attend either to the temporal or the spatial dimension. Here, we use a novel approach, consisting of a cascade of two self-attention mechanisms, one for the temporal and one for the spatial dimension. The proposed strategy first attends to the temporal dimension. The temporal attention output is then fed to the spatial self-attention. As a result, the stacked attentions attend both time and space axes. Mathematically, we define the temporal $\boldsymbol{\alpha}_t$ and spatial $\boldsymbol{\alpha}_s$ self-attention layers \cite{vaswani2017attention} as:

        \begin{equation}
            \boldsymbol{\alpha}_t(\mathbf{x}) = \text{softmax}\left(\frac{(\mathbf{W_{q,t}}\mathbf{x})(\mathbf{W_{k,t}}\mathbf{x})^T}{\sqrt{d_k}}\right)
        \end{equation}

        \begin{equation}
            \boldsymbol{\alpha}_s(\mathbf{x}) = \text{softmax}\left(\frac{(\mathbf{W_{q,s}}\mathbf{x})(\mathbf{W_{k,s}}\mathbf{x})^T}{\sqrt{d_k}}\right)
        \end{equation}

        where $\mathbf{x}$ denotes the generic input to the attention layer. $\mathbf{W_{q,t}}, \mathbf{W_{k,t}}$ and $\mathbf{W_{q,s}}, \mathbf{W_{k,s}}$ are learnable projection matrices for query and key for the temporal and spatial self-attentions, respectively. The output of the temporal self-attention layer is then computed as:

        \begin{equation}
            \mathbf{o}_t = \boldsymbol{\alpha}_t(\mathbf{x}) \mathbf{W_{v,t}}\mathbf{x}
        \end{equation}

        and the output of the stacked self-attentions is then computed as:

        \begin{equation}
            \mathbf{o}_{t,s} = \boldsymbol{\alpha}_s(\mathbf{o}_t)\mathbf{W_{v,s}}\mathbf{o}_t
        \end{equation}

        where $\mathbf{W_{v,t}}, \mathbf{W_{v,s}}$ are learnable projection matrices for the value. We set the output and embedding size of the transformer to 128. After the self-attention, we use a dropout layer to avoid overfitting (dropout rate $\rho=0.1$). We do not derive here the equations of the feed-forward network following the self-attention, nor the normalization layers used inside the self-attention: we refer the reader to the general transformer formulation \cite{vaswani2017attention}. The output of the transformer is then linearly projected using a linear transformation $\mathbf{W_o}$ (fully-connected layer) to output a 2-dimensional (N-S, E-W) denoised time series, obtained from a weighted average of the previous 128 feature maps, modulated, during training, by a dropout layer (dropout rate $\rho=0.5$).

    \subsection{Training details}

        The input of SSEdenoiser is given by the time series $\boldsymbol{\xi}(t)$. We train the model via mini-batch training (batch size of 128) by minimizing the mean squared error (MSE) between the target (clean) $\mathbf{d}(t)$ and the output (denoised) $\mathbf{\hat d}(t)$ time series:

        \begin{align}\label{eq:mse}
        \begin{split}
            \text{MSE}&(\mathbf{d}(t), \mathbf{\hat d}(t)) = \\ & \frac{1}{n \cdot N_s \cdot N_t \cdot N_d} \sum_{i=1}^{n} \sum_{j=1}^{N_s} \sum_{t=1}^{N_t} \sum_{k=1}^{N_d}(d_j^{i,k}(t) - \hat d_j^{i,k}(t))^2
            \end{split}
        \end{align}

        where $n$ is the number of samples in the mini-batch, $N_s$ is the number of stations, $N_t$ is the window length (60 days) and $N_d$ is the number of components (2, N-S and E-W). We choose the MSE as error loss since it is a standard regression error loss and it is effective in penalizing large denoising errors during training. The training database consists of 60,000 samples, which are split into training (80\%), validation (10\%), and test (10\%) sets.

\section{Results} \label{sec:results}

    \subsection{Overall results on synthetic data}\label{sec:overall-results}

        \begin{figure}[!t]
            \centering
            \includegraphics[width=\linewidth]{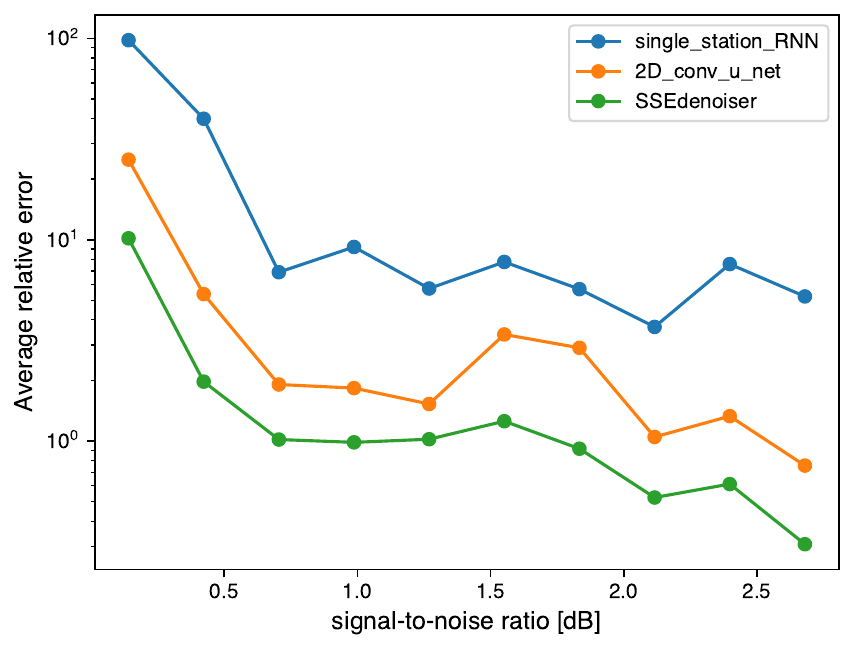}
            \caption{Evaluation of the denoising power as a function of the signal-to-noise ratio for the tested deep-learning models, namely single\_station\_RNN, 2D\_conv\_u\_net and SSEdenoiser. The average absolute error (see equations \ref{eq:denoising-error} and \ref{eq:avg-denoising-error}) for a given SNR bin is plotted.}
            \label{fig:err-vs-snr}
        \end{figure}
        
        We compare SSEdenoiser against both traditional signal processing techniques and deep learning strategies. We adopt moving mean and median filtering as a baseline comparison, with different kernel sizes, here \{3, 7, 15\} days. We will refer to these models as \textbf{moving\_avg\_k} and \textbf{moving\_med\_k}, where \textit{k} indicates the kernel size. Furthermore, we test the following two deep-learning-based methods, which we will refer to as \textbf{single\_station\_RNN} and \textbf{2D\_conv\_u\_net}:

        \begin{itemize}
            \item \textbf{single\_station\_RNN}. This model has been adapted from Xue and Freymueller (2023) \cite{xue2023machine} and it was originally intended as a detector of transient deformation signals in GNSS time series. We modify the last layer to match the size of the input for our denoising application. The model applies a Recurrent Neural Network (RNN) with a Bidirectional Long Short-Term Memory (BiLSTM) cell to each station separately and it is followed by two fully connected layers. We keep the same parameters as the original model, notably 64 for the BiLSTM embedding dimension and 16 for the fully connected layer.
            \item \textbf{2D\_conv\_u\_net}. This model works on GNSS time series stacked to form a matrix representation of dimensions $(N_s, N_t)$ for each component, where the time series are sorted by latitude. The idea is to create a multi-station method working on image-like data, since the rows represent stations that are close in space. To this end, we develop a standard U-Net architecture following its original formulation \cite{ronneberger2015u} and using four blocks (we will refer to this as a convolutional block) with 2D convolution, ReLu activation and Batch Normalization with {[32, 64, 128, 256]} feature maps for the contraction path. We interleave the convolutional blocks with 2D Max Pooling layers. We replace the 2D convolution and the 2D Max Pooling with 2D transposed convolution and upsampling layers for the expanding path, respectively.
        \end{itemize}
        
        We use the mean squared error (MSE) and the mean absolute error (MAE) as error metrics, which we report along with their standard deviation. We use the MSE definition already presented in equation \ref{eq:mse}. We first define the squared error relative to the \textit{i}-th sample as:

        \begin{equation}
            \text{SE}_i(\mathbf{d}(t), \mathbf{\hat d}(t)) = \sum_{j=1}^{N_s} \sum_{t=1}^{N_t} \sum_{k=1}^{N_c} (d_j^{i,k}(t) - \hat d_j^{i,k}(t))^2
        \end{equation}
        
        The corresponding mean and standard deviation are computed as:

        \begin{equation}
            \text{MSE}(\mathbf{d}(t), \mathbf{\hat d}(t)) = \frac{1}{n} \sum_{i=1}^N \text{SE}_i(\mathbf{d}(t), \mathbf{\hat d}(t))
        \end{equation}
        
        \begin{equation}
            \sigma_{SE} = \sqrt{\frac{1}{n} \sum_{i=1}^N \left(\text{SE}_i(\mathbf{d}(t), \mathbf{\hat d}(t)) - \text{MSE}(\mathbf{d}(t), \mathbf{\hat d}(t)) \right)^2}
        \end{equation}

        The absolute error is defined similarly:

        \begin{equation}
            \text{AE}_i(\mathbf{d}(t), \mathbf{\hat d}(t)) = \sum_{j=1}^{N_s} \sum_{t=1}^{N_t} \sum_{k=1}^{N_c} |d_j^{i,k}(t) - \hat d_j^{i,k}(t)|
        \end{equation}

        \begin{equation}
            \text{MAE}(\mathbf{d}(t), \mathbf{\hat d}(t)) = \frac{1}{n} \sum_{i=1}^N \text{AE}_i(\mathbf{d}(t), \mathbf{\hat d}(t))
        \end{equation}
        
        \begin{equation}
            \sigma_{AE} = \sqrt{\frac{1}{n} \sum_{i=1}^N \left(\text{AE}_i(\mathbf{d}(t), \mathbf{\hat d}(t)) - \text{MAE}(\mathbf{d}(t), \mathbf{\hat d}(t)) \right)^2}
        \end{equation}

        \begin{table}
            \caption{Denoising error of the tested methods (mean $\pm$ standard deviation) on the synthetic test set.}
            \centering
            \begin{tabular}{ccc}
                \hline
                Model & $\text{MSE} \pm \sigma_{SE}$  & $\text{MAE} \pm \sigma_{AE}$ \\
                \hline
                moving\_avg\_3 & $5.74 \pm 1.64$ & $1.77 \pm 0.29$   \\
                moving\_med\_3 & $5.85 \pm 1.64$ & $1.78 \pm 0.29$   \\
                moving\_avg\_7 & $5.24 \pm 1.63$ & $1.69 \pm 0.30$   \\
                moving\_med\_7 & $5.15 \pm 1.60$ & $1.65 \pm 0.30$   \\
                moving\_avg\_15 & $4.89 \pm 1.61$ & $1.63 \pm 0.31$   \\
                moving\_med\_15 & $4.50 \pm 1.53$ & $1.51 \pm 0.30$   \\
                single\_station\_RNN & $0.24 \pm 0.34$ & $0.21 \pm 0.15$   \\
                2D\_conv\_u\_net & $0.17 \pm 0.42$ & $0.13 \pm 0.12$   \\
                \textbf{SSEdenoiser} & $\boldsymbol{0.03} \pm \boldsymbol{0.06}$ & $\boldsymbol{0.06} \pm \boldsymbol{0.06}$   \\
                \hline
                \label{table:results}
            \end{tabular}
        \end{table}

        We evaluate the denoising performance of the aforementioned models on a synthetic test set composed of 6,000 samples and we present the numerical results in table \ref{table:results}. Traditional methods (mean and median filtering) are associated with a much higher error than deep-learning-based methods. The error decreases as the kernel size increases, yet not significantly, probably because the GNSS noise cannot be eliminated through simple mean or median filtering. This is particularly true for signals having a low signal-to-noise ratio, which are masked by the noise, where a mean or median filter, especially with a relatively large kernel, would not be able to extract the signal from the noise. We also observe that, when the kernel size increases, the error associated with the median filtering is lower than the mean one, probably because the mean filter has an intrinsic smoothing that increases with the kernel size, failing to preserve the amplitude of the denoised signal. Conversely, deep-learning-based methods are associated with a much lower error, both absolute and mean squared error. Among the deep-learning-based methods, the single-station approach outperforms the traditional methods, yet it is less precise than the multi-station ones (2D\_conv\_u\_net and SSEdenoiser), probably because it cannot distinguish between signals that are registered by multiple stations as well as properly accounting for the spatial variability of the noise. As to the multi-station approaches, SSEdenoiser achieves superior performance, with an average error 10 times lower than 2D\_conv\_u\_net, thanks to the graph representation that helps in leveraging the network geometry as well as intra-station properties emerging from the data. In the 2D\_conv\_u\_net the convolution is applied blindly to stations that are close in latitude and that may not be necessarily more informative than a farther station. Hence, a graph turns out to be more accurate and flexible.

        We also discuss the sensitivity of the models as a function of the signal-to-noise ratio of the input time series. We focus this analysis on the deep-learning models only. First, we define the (average) signal-to-noise ratio for a multi-station setting as the average signal-to-noise ratio over the stations that recorded a displacement for all components:

        \begin{equation}
            \overline{\text{SNR}} = \frac{1}{|\mathcal{S}{'}| \cdot N_c} \sum_{j \in \mathcal{S}{'}}^{|\mathcal{S}{'}|}{\sum_{k=1}^{N_c}{10 \log_{10}\left({\frac{\sum_{t=1}^{N_t}{|\xi_j^k(t)|^2}}{\sum_{t=1}^{N_t}{|n_j^k(t)|^2}}}\right)}}
        \end{equation}

        \begin{equation*}
            \begin{split}
                \mathcal{S}{'} &= \{j: d_j^k(t) \neq 0\}, \\
                &\quad \forall k \in (0, N_c), \forall t \in (0, N_t), \forall j \in (0, N_s)
            \end{split}
        \end{equation*}
        
        where $\mathcal{S}{'}$ indicates a set of indices relative to the stations where a nonzero displacement occurred. Hence, stations that did not record any displacement are not taken into account. To compare the denoising performance, we define the denoising error as the mean absolute error relative to the maximum amplitude displacement for the \textit{i}-th sample:

        \begin{equation}
            \mathcal{E}(i) = \frac{1}{N_t \cdot N^{'}_s \cdot N_c} \sum_{j \in \mathcal{S}{'}}^{|\mathcal{S}{'}|}{\sum_{k=1}^{N_c}{\frac{\sum_{t=1}^{N_t}{|d_j^{i, k}(t) - \hat d_j^{i, k}(t)|}}{\max_t{|d_j^{i, k}(t)|}}}}
            \label{eq:denoising-error}
        \end{equation}

        and we compute the average denoising error as:

        \begin{equation}
            \Bar{\mathcal{E}} = \frac{1}{n} \sum_{i=1}^n {\mathcal{E}(i)}
            \label{eq:avg-denoising-error}
        \end{equation}

        This allows us to compare the denoising power of the models regardless of the displacement amplitude, so that statistics can be made. Figure \ref{fig:err-vs-snr} shows the average denoising error as a function of the (binned) signal-to-noise ratio for the three deep-learning-based models. In all models, the average error naturally decreases as the signal-to-noise ratio increases. Globally, SSEdenoiser performs the best among the three models, as also seen in table \ref{table:results}. For low values of signal-to-noise ratio (\textit{e.g.}, $\overline{\text{SNR}}<1$), the three models behave differently and the performance is much more degraded for the single\_station\_RNN model, with an average error on the displacement which is 1.3 times higher than the one obtained for $\overline{\text{SNR}}>1$. The other two models, namely 2D\_conv\_u\_net and SSEdenoiser, are more stable. For $\overline{\text{SNR}}<1$, their average error is 0.4 and 0.1 times larger than for $\overline{\text{SNR}}>1$, respectively. SSEdenoiser exhibits better performance and stability for all values of $\overline{\text{SNR}}$, with the average error for $\overline{\text{SNR}}=0$ being 4 times lower than 2D\_conv\_u\_net and 12 times lower than single\_station\_RNN.

    \subsection{Ablation study}

    \begin{table}
            \caption{Denoising error (mean $\pm$ standard deviation) on the synthetic test set for the ablation study.}
            \centering
            \begin{tabular}{ccc}
                \hline
                Model & $\text{MSE} \pm \sigma_{SE}$  & $\text{MAE} \pm \sigma_{AE}$ \\
                \hline
                no\_transformer & $0.05 \pm 0.07$ & $0.10 \pm 0.06$   \\
                spatial\_attention\_only & $0.04 \pm 0.07$ & $0.08 \pm 0.06$   \\
                temporal\_attention\_only & $\boldsymbol{0.03} \pm \boldsymbol{0.06}$ & $0.07 \pm 0.06$   \\
                \hline
                \textbf{SSEdenoiser} & $\boldsymbol{0.03} \pm \boldsymbol{0.06}$ & $\boldsymbol{0.06} \pm \boldsymbol{0.06}$   \\
                \hline
                \label{table:ablation}
            \end{tabular}
        \end{table}

        \begin{figure}[!t]
            \centering
            \includegraphics[width=\linewidth]{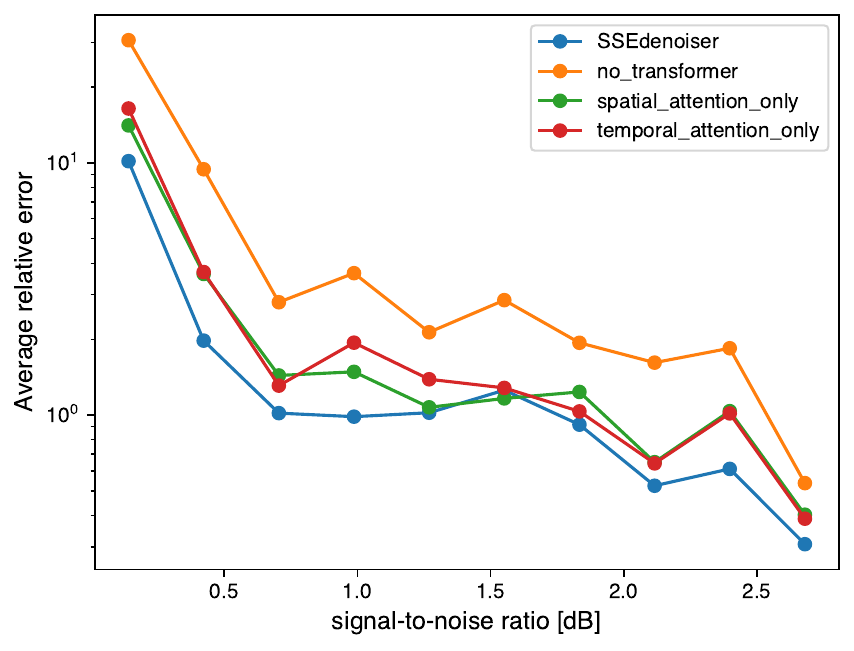}
            \caption{Evaluation of the denoising power as a function of the signal-to-noise ratio for the ablated models, namely no\_transformer, spatial\_attention\_only and temporal\_attention\_only, against SSEdenoiser. The average absolute error for a given SNR bin is plotted.}
            \label{fig:ablation-err-vs-snr}
        \end{figure}
        
        The novelty of SSEdenoiser lies in the combination of a graph-based RNN \cite{bai2020adaptive} with a spatiotemporal transformer module. We thus perform an ablation study to test the robustness of the proposed spatiotemporal transformer. We test SSEdenoiser against three ablated models, obtained by completely removing the transformer, by using temporal and spatial attention mechanisms separately, namely \textbf{no\_transformer}, \textbf{spatial\_attention\_only} and \textbf{temporal\_attention\_only}. We train and test each ablated model with the same configuration as SSEdenoiser and we report the results in table \ref{table:ablation}.

        Removing the transformer results in a higher global error, almost two times higher than that of SSEdenoiser. When adding the spatial attention, the global error decreases, with the model incorporating the temporal attention corresponding to the lowest misfit among the ablated models. Compared to SSEdenoiser, the temporal attention configuration exhibits the same squared error, yet a higher absolute error.

        We further compare the four models by evaluating the denoising power (similarly to what is done in section \ref{sec:overall-results}), by computing the average relative error as a function of the signal-to-noise ratio. The \textbf{no\_transformer} model exhibits lower performance for any value of SNR, further motivating the need for incorporating an attention mechanism. The \textbf{spatial\_attention\_only} and \textbf{temporal\_attention\_only} have almost equivalent performance. The combination of both spatial and temporal attention (SSEdenoiser) results in higher performance at almost every scale of SNR, showing that, regardless of the noise level, the proposed model can have superior denoising power by coupling attention on the spatial and temporal axis, which cannot be decoupled without resulting in a loss of resolution.
    
    \subsection{Results on synthetic data: learned station connectivity}

        \begin{figure*}[!t]
            \centering
            \includegraphics[width=\linewidth]{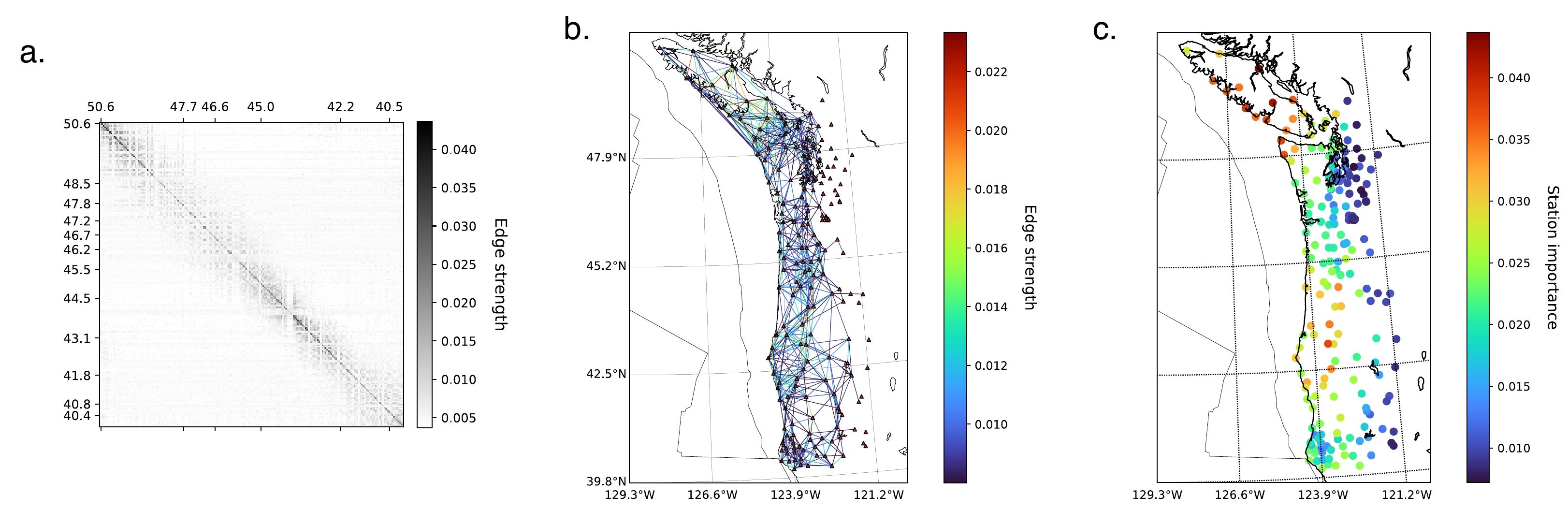}
            \caption{Graph connectivity learned by SSEdenoiser. (a) learned adjacency matrix. Nodes are sorted by latitude and color-coded by the learned edge strength. (b) Geographic representation of the strongest connections (edge strength values in the range (0.008, 0.0234) color-coded by edge strength). (c) GNSS stations used in this study, color-coded by the learned station importance, \textit{i.e.}, the value of the diagonal of the adjacency matrix for each node.}
            \label{fig:learned-connectivity}
        \end{figure*}
        
        Our multi-station approach also allows access to spatial and temporal relationships between GNSS stations for further interpretation. During training, SSEdenoiser learns which stations (nodes of the graph) should be connected and how strong their connections should be: this information is synthesized in the adjacency matrix, that provides the edge strength of the graph. 
        In Figure \ref{fig:learned-connectivity}(a), we show the learned adjacency matrix, with nodes sorted by latitude (each pixel represents a relationship between two stations). The number of edges in the graph is $N + \frac{N (N-1)}{2} = 20100$ including self-loops. We see that SSEdenoiser has learned to connect nodes that are mostly spatially close to each other (near the diagonal), yet it also allows for weaker long-range connections (\textit{e.g.}, they are assigned a lower edge strength). This suggests that the method could generally rely on information available within a neighborhood and then compare information coming from different sub-networks. Moreover, this information is entirely learned from the data, since the model is not fed with any explicit information about the station location.

        To visualize the backbone of the graph structure, we further filter the adjacency matrix by selecting edge strength values higher than 0.008, corresponding to the connections with strong edge weights. We found 878 connections (excluding the self-connections on the diagonal) to which we will refer in the following as ``strong connections''. We show their spatial distribution between GNSS stations in Figure \ref{fig:learned-connectivity}(b). We first see that these connections are such that the azimuthal coverage is as high as possible between neighbouring stations. Also, the method has learned how to produce a mesh connecting all the stations that are located on top of the slow slip area used in the training phase (between 20 and 40 km depth, see Figure \ref{fig:data}(b)). Stations located further inland (longitude $< 122$°) do not have strong edge weights, suggesting that they are not very informative for slow slip detection given the location of the SSE area. Also, the stronger edges connect stations that are in areas where the station coverage is sparse. When the network is dense, the high weights are indeed less useful because the information can already be included in the signal coming from the numerous nearby stations. 
    
        Figure \ref{fig:learned-connectivity}(c) shows the 200 GNSS stations color-coded by the value of the adjacency matrix diagonal term. It indicates the strength of the self-loop connections for each station, which can be thought of as a measure of the learned station's self-relevance (or importance). We can see that the importance of further inland stations is low compared to stations that are located above the SSE area. The highest importance values are assigned to stations located in Vancouver Island and at latitudes between 42.5° and 45.2°, probably linked to a learned trade-off between slow slip occurrence and coverage of GNSS stations. This can be seen as a proxy of the distance to the SSE source area combined with the local density of stations.

    \subsection{Denoising of real non-post-processed GNSS time series in Cascadia in 2007-2022} 

        \begin{figure*}[!t]
            \centering
            \includegraphics[width=\linewidth]{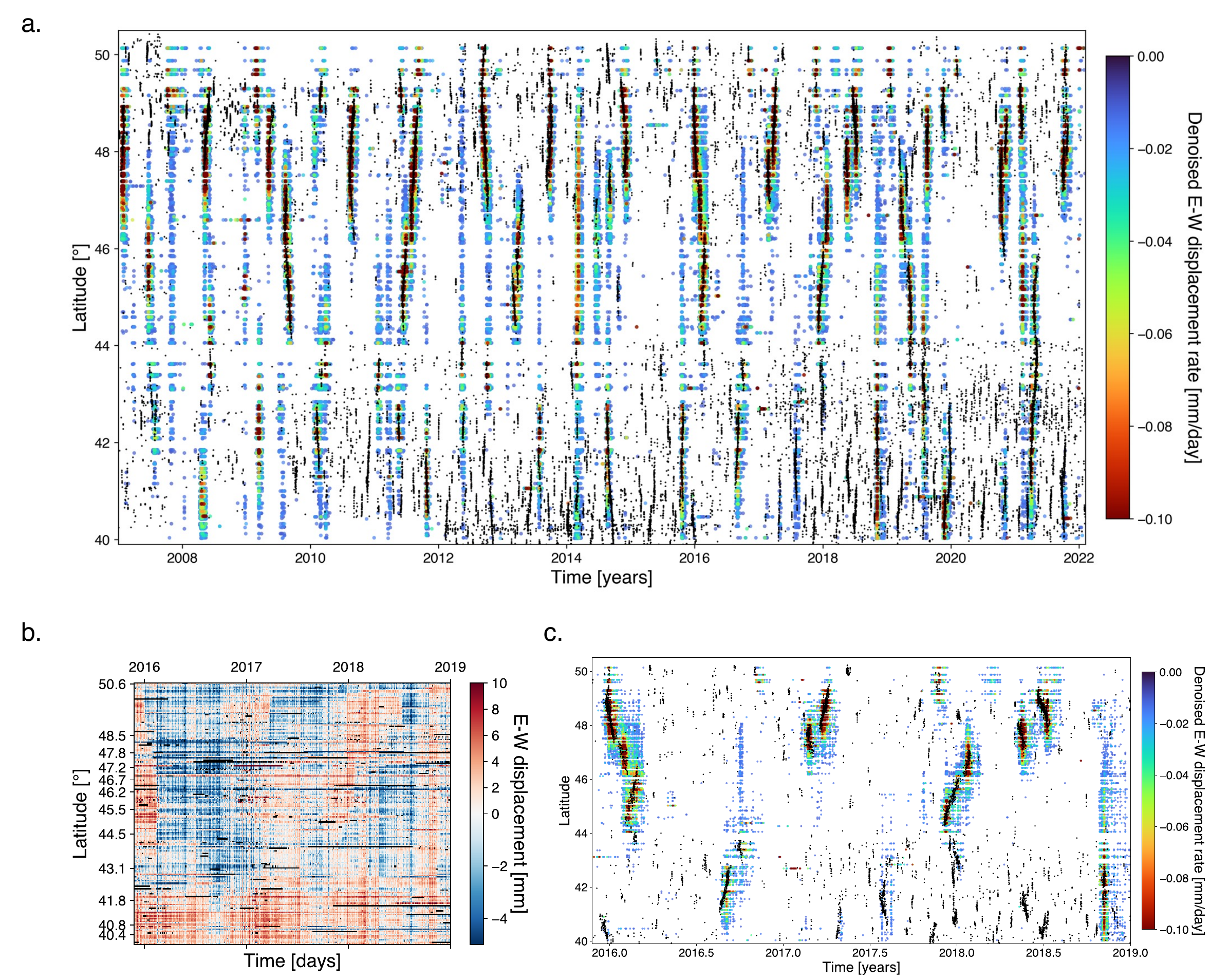}
            \caption{Denoising of real GNSS time series in Cascadia from 2007 to 2022. (a) Denoised displacement rate (E-W component) as a function of time. The displacement rate computed from the output of SSEdenoiser is shown for each GNSS station, sorted as a function of the latitude. Tremor epicenters are also shown as black points (we use the catalogue from Ide, 2012 \cite{ide2012variety} until August 5, 2009, and the tremor catalogue from the Pacific Northwest Seismic Network (PNSN) in the following period \cite{wech2008automated}). (b) Example of raw GNSS data for the period 2016-2019. Each row of the matrix represents a detrended GNSS time series, color-coded by the amount of displacement in the E-W component. Stations are sorted by latitude. (c)Denoised displacement field computed from the output of SSEdenoiser from the same input time series as in the (b) panel. The displacement rate is shown by the colored points. The black points represent tremor epicenters. Denoised displacements are very well correlated with the occurrence of tremors, which is an expected feature that provides independent validation of the results.}
            \label{fig:denoising-real-data}
        \end{figure*}
        
        We test SSEdenoiser on real non-post-processed GNSS time series from 2007 to 2022. We take 60-day windows of data and apply SSEdenoiser on each of them, by sliding the window with a stride of 1 day and collecting the resulting denoised GNSS time series. In practice, non-overlapping sliding windows can be used (stride of 60 days). Here, we prefer relying on multiple denoising outputs (stride of 1 day) to provide a better estimate of the displacement by averaging the contributions coming from all the possible sliding windows. We compute the temporal derivative of the estimated denoised time series in each window to obtain the displacement rate and we keep the 20 days in the middle of the window to exclude potential border effects. For each time step, we take the mean of all 20-day overlapping windows (20 windows). With this procedure, we obtain the daily average displacement rate.

        Figure \ref{fig:denoising-real-data}(a) shows the obtained denoised GNSS time series of daily (E-W) displacement rates at all stations in matrix form, over the period 2007-2022. For simplicity, only displacement rates larger than 0.01 mm/day are shown. The retrieved displacement rate has a coherent spatiotemporal distribution: it occurs by bursts that are clustered in latitude and time, consistent with previous studies \cite{schmidt2010source, michel2019interseismic}. The amplitude is nearly always negative (meaning a displacement in the west direction), that is the direction of expected motion. The largest slow slip events are associated with large displacement rates, such as the May 2011 or the January 2016 slow slip event (see zoom in Figure \ref{fig:denoising-real-data}(c)). We can see that SSEdenoiser is also able to constrain slow slip occurring in South Cascadia, which is more difficult than in the northern area because of the known higher noise level.
        
        The denoised displacement rates have a good correlation with the spatiotemporal distribution of tremors (shown in black in the figures). Tremors are low-frequency and low-amplitude seismic events that have been observed to accompany slow slip events in Cascadia, \textit{e.g.} \cite{rogers2003episodic}. SSEdenoiser is blindly trained on GNSS time series, without incorporating any information from tremors in the model: this means that our method is validated thanks to this independent information. The displacement distribution follows the tremor propagation in space and time, both for large and smaller slow slip events as well as for propagating events, as can be seen in the zoom in Figure \ref{fig:denoising-real-data}(c). Our method can effectively retrieve slip migration, such as in the case of the 2016 or 2018 slow slip events as well as recognizing events that are close in time and space, such as the Feb.-Apr. 2017 or June-Aug. 2018 events. This suggests that SSEdenoiser has effectively learned what the noise structure looks like to retrieve the concealed slow deformation at any space and time scale, especially by comparing the denoised displacement with the raw data that was fed in input to the model, as we see in Figure \ref{fig:denoising-real-data}(b).
        

    \subsection{Comparison between SSEdenoiser and the single-station method by Xue and Freymueller (2023)}

        \begin{figure*}[!t]
            \centering
            \includegraphics[width=\linewidth]{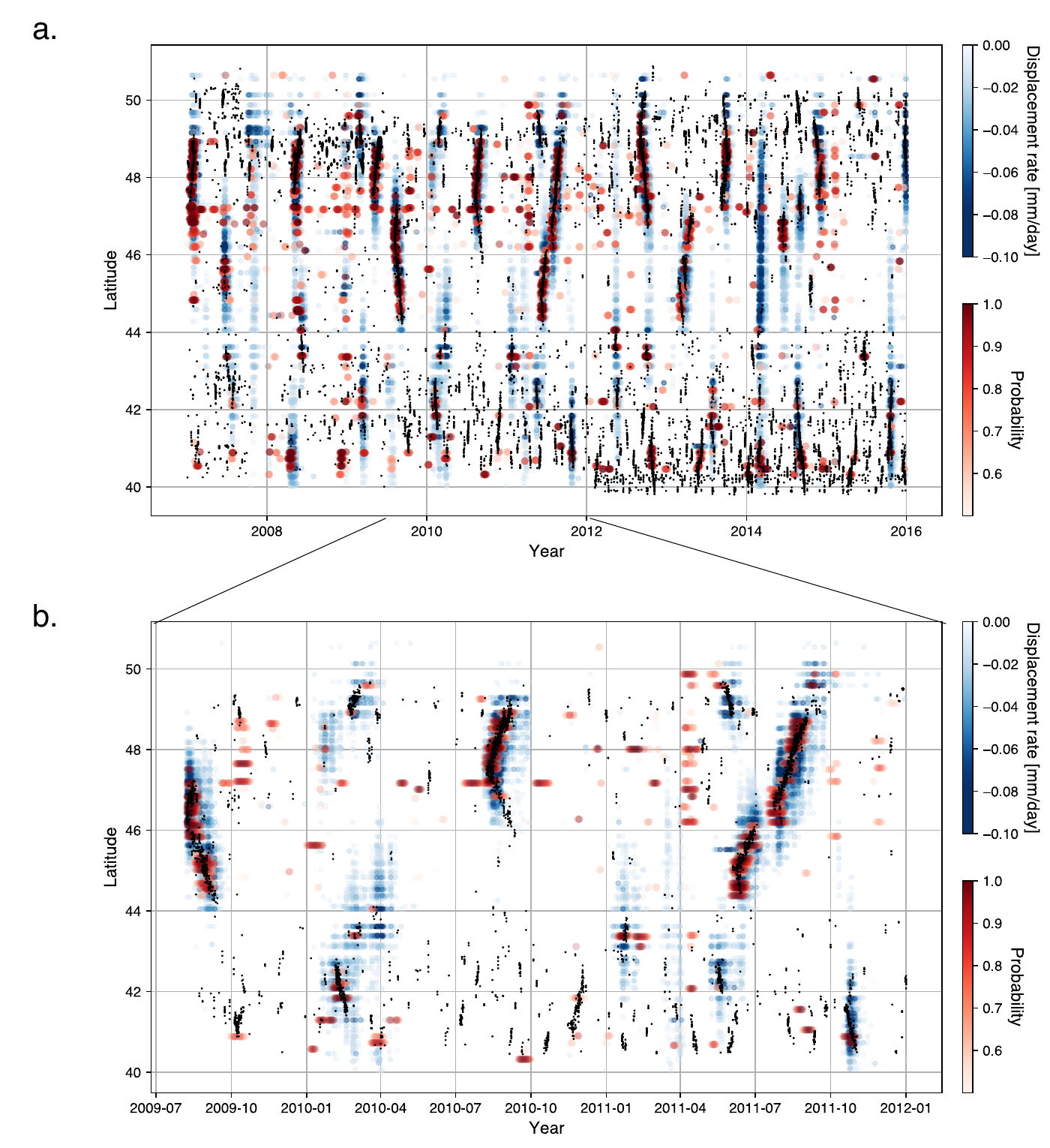}
            \caption{(a) Qualitative comparison between SSEdenoiser and the single-station method by Xue and Freymueller (2023) \cite{xue2023machine} in Cascadia from 2007 to 2016. Black points represent tremor epicenters, sorted by latitude. Red points are associated with the probability of SSE occurrence by Xue and Freymueller \cite{xue2023machine} and blue points represent the denoised displacement field obtained by running SSEdetector on the raw data in Cascadia from 2007 to 2016. (b) Zoom in the period ranging from Aug. 2009 to Dec. 2011.}
            \label{fig:costantino-vs-xue}
        \end{figure*}
        
        We qualitatively compare the denoised displacements computed from the output of SSEdenoiser with the results obtained by Xue and Freymueller (2023) \cite{xue2023machine}. They developed a single-station deep learning method (cf. section \ref{sec:overall-results})) whose output is the probability of occurrence of a slow slip event for each separate GNSS time series. Figure \ref{fig:costantino-vs-xue}(a) shows the SSEdenoiser denoised displacements along with this SSE probability. There is, globally, good accordance between our denoised time series and the SSE probability obtained by Xue et Freymueller, which can be thought of as a proxy of the SSE-driven displacement.

        Xue and Freymueller's prediction probability (in red) has a good correlation with tremor epicenters, yet it exhibits some level of noise and inconsistencies along the stations. For instance, the horizontal cluster of points at latitude 47° seems to suggest that, in the case of perturbed time series (which seems likely the case), the ability to discern coherence in the signal is not properly constrained due to the absence of corroborative data from other stations. Also, another limitation inherent in single-station analysis is that these models tend to produce isolated points, as we see in Figure \ref{fig:costantino-vs-xue}. These isolated points are associated with a high probability, yet there is no agreement from neighbouring stations. Hence, these can be likely considered false positives.
        
        Conversely, SSEdenoiser gains in accuracy thanks to the multi-station approach. The denoised displacement output by SSEdenoiser follows the tremor distribution in space and time, suggesting that our approach has the potential to provide finer-scale observations of slow slip by leveraging the spatial correlations between the GNSS time series as well as the GNSS geometry. Figure \ref{fig:costantino-vs-xue}(b) shows a zoom in the period ranging from August 2009 to December 2011. Our method seems to show a better ability to constrain the spatial consistency of the SSE-driven displacement, such as in the case of big events, \textit{e.g.}, the September 2009 slow slip event, where the probability of the Xue and Freymueller's approach does not exceed the detection threshold for all the stations at latitudes between 47° and 48°, as opposed to the displacement retrieved by SSEdenoiser, consistent with tremor recordings. The same seems to happen also for smaller events, \textit{e.g.}, the November 2011 slow slip event (average latitude of 41°) and the June 2011 event (latitude 47°), where our method shows the potential of revealing slow slip with improved detail thanks to a multi-station approach. Yet, SSEdenoiser also has some limitations, such as the vertical clusters (\textit{e.g.}, early 2014 or late 2018), which can be ascribed to the complexity of the synthetic training set. These vertical clusters are most probably the signature of GNSS common mode errors in the real data, which might not be perfectly modeled by our noise generation strategy and need further improvement.

\section{Conclusion}

    We develop SSEdenoiser, a graph-based deep learning method for denoising non-post-processed GNSS position time series. We build a synthetic database, consisting of realistic noise and synthetic slow slip event signals. We first compare SSEdenoiser to traditional denoising methods as well as single- and multi-station-based deep learning models. We find that our method has superior accuracy and stability also for very low values of signal-to-noise ratio. We also analyze the characteristics of SSEdenoiser by looking at the learned adjacency matrix. We find that SSEdenoiser learns to connect stations based on the region where SSEs are located in the training samples and to the density of GNSS stations.
    
    When tested on real data, SSEdenoiser proves effective in isolating the displacement related to slow slip events, with remarkable spatial and temporal correlation with tremors, which are not given as input to the method. By using tremors as independent validation, we conclude that our method can denoise GNSS data and retrieve concealed transient deformation at all scales of both duration and signal amplitude.
    
    We finally perform a qualitative test between SSEdenoiser and the single-station method by Xue and Freymueller (2023) \cite{xue2023machine} and we find that SSEdenoiser exhibits superior performance in terms of spatiotemporal distribution of the denoised displacement as well as consistency with tremor episodes in space and time.
    
    Our approach also has the advantage that denoised time series could be used as input for traditional slip inversion techniques as well as to better constrain the source parameters. Since the level of noise in GNSS time series is such that regularization constraints must be used to guarantee temporal smoothness, our denoised time series could be also used for inverting the slip on the subduction interface more easily than the non-denoised data.

\section*{Acknowledgments}
This work has been funded by ERC CoG 865963 DEEP-trigger and by MIAI@Grenoble Alpes (ANR-19-P3IA-0003). The deep learning model training was performed using HPC resources from GENCI-IDRIS (Grant 2023-AD011012373R2).



\bibliographystyle{IEEEtran}
\bibliography{bibtex/bib/IEEEabrv, bibtex/bib/IEEEexample}

\begin{thebibliography}{10}
\providecommand{\url}[1]{#1}
\csname url@samestyle\endcsname
\providecommand{\newblock}{\relax}
\providecommand{\bibinfo}[2]{#2}
\providecommand{\BIBentrySTDinterwordspacing}{\spaceskip=0pt\relax}
\providecommand{\BIBentryALTinterwordstretchfactor}{4}
\providecommand{\BIBentryALTinterwordspacing}{\spaceskip=\fontdimen2\font plus
\BIBentryALTinterwordstretchfactor\fontdimen3\font minus \fontdimen4\font\relax}
\providecommand{\BIBforeignlanguage}[2]{{%
\expandafter\ifx\csname l@#1\endcsname\relax
\typeout{** WARNING: IEEEtran.bst: No hyphenation pattern has been}%
\typeout{** loaded for the language `#1'. Using the pattern for}%
\typeout{** the default language instead.}%
\else
\language=\csname l@#1\endcsname
\fi
#2}}
\providecommand{\BIBdecl}{\relax}
\BIBdecl

\bibitem{luo2023multiscale}
F.~Luo, T.~Zhou, J.~Liu, T.~Guo, X.~Gong, and J.~Ren, ``Multiscale diff-changed feature fusion network for hyperspectral image change detection,'' \emph{IEEE Transactions on Geoscience and Remote Sensing}, vol.~61, pp. 1--13, 2023.

\bibitem{hong2007experimental}
Y.~Hong, R.~F. Adler, and G.~Huffman, ``An experimental global prediction system for rainfall-triggered landslides using satellite remote sensing and geospatial datasets,'' \emph{IEEE Transactions on Geoscience and Remote Sensing}, vol.~45, no.~6, pp. 1671--1680, 2007.

\bibitem{yi2020new}
Y.~Yi and W.~Zhang, ``A new deep-learning-based approach for earthquake-triggered landslide detection from single-temporal rapideye satellite imagery,'' \emph{IEEE Journal of Selected Topics in Applied Earth Observations and Remote Sensing}, vol.~13, pp. 6166--6176, 2020.

\bibitem{wagner2012geospatial}
M.~A. Wagner, S.~W. Myint, and R.~S. Cerveny, ``Geospatial assessment of recovery rates following a tornado disaster,'' \emph{IEEE transactions on geoscience and remote sensing}, vol.~50, no.~11, pp. 4313--4322, 2012.

\bibitem{ding2022bi}
L.~Ding, H.~Guo, S.~Liu, L.~Mou, J.~Zhang, and L.~Bruzzone, ``Bi-temporal semantic reasoning for the semantic change detection in hr remote sensing images,'' \emph{IEEE Transactions on Geoscience and Remote Sensing}, vol.~60, pp. 1--14, 2022.

\bibitem{benediktsson2005classification}
J.~A. Benediktsson, J.~A. Palmason, and J.~R. Sveinsson, ``Classification of hyperspectral data from urban areas based on extended morphological profiles,'' \emph{IEEE Transactions on Geoscience and Remote Sensing}, vol.~43, no.~3, pp. 480--491, 2005.

\bibitem{ghamisi2014fusion}
P.~Ghamisi, J.~A. Benediktsson, and S.~Phinn, ``Fusion of hyperspectral and lidar data in classification of urban areas,'' in \emph{2014 IEEE Geoscience and Remote Sensing Symposium}.\hskip 1em plus 0.5em minus 0.4em\relax IEEE, 2014, pp. 181--184.

\bibitem{camps2016sensitivity}
A.~Camps, H.~Park, M.~Pablos, G.~Foti, C.~P. Gommenginger, P.-W. Liu, and J.~Judge, ``Sensitivity of gnss-r spaceborne observations to soil moisture and vegetation,'' \emph{IEEE Journal of Selected Topics in Applied Earth Observations and Remote Sensing}, vol.~9, no.~10, pp. 4730--4742, 2016.

\bibitem{pierdicca2021potential}
N.~Pierdicca, D.~Comite, A.~Camps, H.~Carreno-Luengo, L.~Cenci, M.~P. Clarizia, F.~Costantini, L.~Dente, L.~Guerriero, A.~Mollfulleda \emph{et~al.}, ``The potential of spaceborne gnss reflectometry for soil moisture, biomass, and freeze--thaw monitoring: Summary of a european space agency-funded study,'' \emph{IEEE geoscience and remote sensing magazine}, vol.~10, no.~2, pp. 8--38, 2021.

\bibitem{dai2020entering}
K.~Dai, Z.~Li, Q.~Xu, R.~B{\"u}rgmann, D.~G. Milledge, R.~Tomas, X.~Fan, C.~Zhao, X.~Liu, J.~Peng \emph{et~al.}, ``Entering the era of earth observation-based landslide warning systems: A novel and exciting framework,'' \emph{IEEE Geoscience and Remote Sensing Magazine}, vol.~8, no.~1, pp. 136--153, 2020.

\bibitem{unwin2021introduction}
M.~J. Unwin, N.~Pierdicca, E.~Cardellach, K.~Rautiainen, G.~Foti, P.~Blunt, L.~Guerriero, E.~Santi, and M.~Tossaint, ``An introduction to the hydrognss gnss reflectometry remote sensing mission,'' \emph{IEEE Journal of Selected Topics in Applied Earth Observations and Remote Sensing}, vol.~14, pp. 6987--6999, 2021.

\bibitem{garrison2014ieee}
J.~L. Garrison and E.~Cardellach, ``The ieee gnss and signals of opportunity working group [technical committees],'' \emph{IEEE Geoscience and Remote Sensing Magazine}, vol.~2, no.~4, pp. 54--58, 2014.

\bibitem{zavorotny2014tutorial}
V.~U. Zavorotny, S.~Gleason, E.~Cardellach, and A.~Camps, ``Tutorial on remote sensing using gnss bistatic radar of opportunity,'' \emph{IEEE Geoscience and Remote Sensing Magazine}, vol.~2, no.~4, pp. 8--45, 2014.

\bibitem{li2014high}
X.~Li, M.~Ge, C.~Lu, Y.~Zhang, R.~Wang, J.~Wickert, and H.~Schuh, ``High-rate gps seismology using real-time precise point positioning with ambiguity resolution,'' \emph{IEEE Transactions on Geoscience and Remote Sensing}, vol.~52, no.~10, pp. 6165--6180, 2014.

\bibitem{montillet2012extracting}
J.-P. Montillet, P.~Tregoning, S.~McClusky, and K.~Yu, ``Extracting white noise statistics in gps coordinate time series,'' \emph{IEEE Geoscience and Remote Sensing Letters}, vol.~10, no.~3, pp. 563--567, 2012.

\bibitem{alregib2018subsurface}
G.~AlRegib, S.~Fomel, and R.~Lopes, ``Subsurface exploration: recent advances in geo-signal processing, interpretation, and learning [from the guest editors],'' \emph{IEEE Signal Processing Magazine}, vol.~35, no.~2, pp. 16--18, 2018.

\bibitem{malfante2018machine}
M.~Malfante, M.~Dalla~Mura, J.-P. M{\'e}taxian, J.~I. Mars, O.~Macedo, and A.~Inza, ``Machine learning for volcano-seismic signals: Challenges and perspectives,'' \emph{IEEE Signal Processing Magazine}, vol.~35, no.~2, pp. 20--30, 2018.

\bibitem{peixoto2021tensor}
A.~A.~T. Peixoto, C.~A.~R. Fernandes, P.~E.~E. Lara, A.~Inza, J.~I. Mars, J.-P. M{\'e}taxian, M.~Dalla~Mura, and M.~Malfante, ``Tensor-based learning framework for automatic multichannel volcano-seismic classification,'' \emph{IEEE Journal of Selected Topics in Applied Earth Observations and Remote Sensing}, vol.~14, pp. 4517--4529, 2021.

\bibitem{lara2020automatic}
P.~E.~E. Lara, C.~A.~R. Fernandes, A.~Inza, J.~I. Mars, J.-P. M{\'e}taxian, M.~Dalla~Mura, and M.~Malfante, ``Automatic multichannel volcano-seismic classification using machine learning and emd,'' \emph{IEEE Journal of Selected Topics in Applied Earth Observations and Remote Sensing}, vol.~13, pp. 1322--1331, 2020.

\bibitem{lopez2020contribution}
M.~L{\'o}pez-P{\'e}rez, L.~Garc{\'\i}a, C.~Ben{\'\i}tez, and R.~Molina, ``A contribution to deep learning approaches for automatic classification of volcano-seismic events: deep gaussian processes,'' \emph{IEEE Transactions on Geoscience and Remote Sensing}, vol.~59, no.~5, pp. 3875--3890, 2020.

\bibitem{chai2020deep}
X.~Chai, G.~Tang, S.~Wang, K.~Lin, and R.~Peng, ``Deep learning for irregularly and regularly missing 3-d data reconstruction,'' \emph{IEEE Transactions on Geoscience and Remote Sensing}, vol.~59, no.~7, pp. 6244--6265, 2020.

\bibitem{zhang2021deep}
W.~Zhang and J.~Gao, ``Deep-learning full-waveform inversion using seismic migration images,'' \emph{IEEE Transactions on Geoscience and Remote Sensing}, vol.~60, pp. 1--18, 2021.

\bibitem{lin2023hybrid}
X.~Lin, C.~Xu, G.~Jiang, and J.~Zang, ``A hybrid deep learning model for rapid probabilistic earthquake source parameter estimation with displacement waveforms from a flexible set of seismic or hr-gnss stations,'' \emph{IEEE Transactions on Geoscience and Remote Sensing}, vol.~61, pp. 1--16, 2023.

\bibitem{shahvandi2021modified}
M.~K. Shahvandi and B.~Soja, ``Modified deep transformers for gnss time series prediction,'' in \emph{2021 IEEE International Geoscience and Remote Sensing Symposium IGARSS}.\hskip 1em plus 0.5em minus 0.4em\relax IEEE, 2021, pp. 8313--8316.

\bibitem{enge1994global}
P.~K. Enge, ``The global positioning system: Signals, measurements, and performance,'' \emph{International Journal of Wireless Information Networks}, vol.~1, pp. 83--105, 1994.

\bibitem{wdowinski1997southern}
S.~Wdowinski, Y.~Bock, J.~Zhang, P.~Fang, and J.~Genrich, ``Southern california permanent gps geodetic array: Spatial filtering of daily positions for estimating coseismic and postseismic displacements induced by the 1992 landers earthquake,'' \emph{Journal of Geophysical Research: Solid Earth}, vol. 102, no.~B8, pp. 18\,057--18\,070, 1997.

\bibitem{williams2004error}
S.~D. Williams, Y.~Bock, P.~Fang, P.~Jamason, R.~M. Nikolaidis, L.~Prawirodirdjo, M.~Miller, and D.~J. Johnson, ``Error analysis of continuous gps position time series,'' \emph{Journal of Geophysical Research: Solid Earth}, vol. 109, no.~B3, 2004.

\bibitem{marill2021fourteen}
L.~Marill, D.~Marsan, A.~Socquet, M.~Radiguet, N.~Cotte, and B.~Rousset, ``Fourteen-year acceleration along the japan trench,'' \emph{Journal of Geophysical Research: Solid Earth}, vol. 126, no.~11, p. e2020JB021226, 2021.

\bibitem{bedford2018greedy}
J.~Bedford and M.~Bevis, ``Greedy automatic signal decomposition and its application to daily gps time series,'' \emph{Journal of Geophysical Research: Solid Earth}, vol. 123, no.~8, pp. 6992--7003, 2018.

\bibitem{michel2019interseismic}
S.~Michel, A.~Gualandi, and J.-P. Avouac, ``Interseismic coupling and slow slip events on the cascadia megathrust,'' \emph{Pure and Applied Geophysics}, vol. 176, no.~9, pp. 3867--3891, 2019.

\bibitem{saad2020deep}
O.~M. Saad and Y.~Chen, ``Deep denoising autoencoder for seismic random noise attenuation,'' \emph{Geophysics}, vol.~85, no.~4, pp. V367--V376, 2020.

\bibitem{zhu2019seismic}
W.~Zhu, S.~M. Mousavi, and G.~C. Beroza, ``Seismic signal denoising and decomposition using deep neural networks,'' \emph{IEEE Transactions on Geoscience and Remote Sensing}, vol.~57, no.~11, pp. 9476--9488, 2019.

\bibitem{thomas2023deep}
A.~Thomas, D.~Melgar, S.~N. Dybing, and J.~R. Searcy, ``Deep learning for denoising high-rate global navigation satellite system data,'' \emph{Seismica}, vol.~2, no.~1, 2023.

\bibitem{munchmeyer2021earthquake}
J.~M{\"u}nchmeyer, D.~Bindi, U.~Leser, and F.~Tilmann, ``Earthquake magnitude and location estimation from real time seismic waveforms with a transformer network,'' \emph{Geophysical Journal International}, vol. 226, no.~2, pp. 1086--1104, 2021.

\bibitem{van2020automated}
M.~P. van~den Ende and J.-P. Ampuero, ``Automated seismic source characterization using deep graph neural networks,'' \emph{Geophysical Research Letters}, vol.~47, no.~17, p. e2020GL088690, 2020.

\bibitem{costantino2023seismic}
\BIBentryALTinterwordspacing
G.~Costantino, S.~Giffard-Roisin, D.~Marsan, L.~Marill, M.~Radiguet, M.~D. Mura, G.~Janex, and A.~Socquet, ``Seismic source characterization from gnss data using deep learning,'' \emph{Journal of Geophysical Research: Solid Earth}, vol. 128, no.~4, p. e2022JB024930, 2023. [Online]. Available: \url{https://agupubs.onlinelibrary.wiley.com/doi/abs/10.1029/2022JB024930}
\BIBentrySTDinterwordspacing

\bibitem{costantino2023multi}
G.~Costantino, S.~Giffard-Roisin, M.~Radiguet, M.~Dalla~Mura, D.~Marsan, and A.~Socquet, ``Multi-station deep learning on geodetic time series detects slow slip events in cascadia,'' \emph{Communications Earth \& Environment}, vol.~4, no.~1, p. 435, 2023.

\bibitem{yu2017spatio}
B.~Yu \emph{et~al.}, ``Spatio-temporal graph convolutional networks: A deep learning framework for traffic forecasting,'' \emph{arXiv preprint arXiv:1709.04875}, 2017.

\bibitem{bai2020adaptive}
L.~Bai \emph{et~al.}, ``Adaptive graph convolutional recurrent network for traffic forecasting,'' \emph{Advances in neural information processing systems}, 2020.

\bibitem{shi2019two}
L.~Shi \emph{et~al.}, ``Two-stream adaptive graph convolutional networks for skeleton-based action recognition,'' in \emph{Proceedings of the IEEE/CVF conference on computer vision and pattern recognition}, 2019, pp. 12\,026--12\,035.

\bibitem{okada1985surface}
Y.~Okada, ``Surface deformation due to shear and tensile faults in a half-space,'' \emph{Bulletin of the seismological society of America}, vol.~75, no.~4, pp. 1135--1154, 1985.

\bibitem{vaswani2017attention}
A.~Vaswani, N.~Shazeer, N.~Parmar, J.~Uszkoreit, L.~Jones, A.~N. Gomez, {\L}.~Kaiser, and I.~Polosukhin, ``Attention is all you need,'' \emph{Advances in neural information processing systems}, vol.~30, 2017.

\bibitem{xue2023machine}
X.~Xue and J.~T. Freymueller, ``Machine learning for single-station detection of transient deformation in gps time series with a case study of cascadia slow slip,'' \emph{Journal of Geophysical Research: Solid Earth}, vol. 128, no.~2, p. e2022JB024859, 2023.

\bibitem{ronneberger2015u}
O.~Ronneberger, P.~Fischer, and T.~Brox, ``U-net: Convolutional networks for biomedical image segmentation,'' in \emph{Medical Image Computing and Computer-Assisted Intervention--MICCAI 2015: 18th International Conference, Munich, Germany, October 5-9, 2015, Proceedings, Part III 18}.\hskip 1em plus 0.5em minus 0.4em\relax Springer, 2015, pp. 234--241.

\bibitem{ide2012variety}
S.~Ide, ``Variety and spatial heterogeneity of tectonic tremor worldwide,'' \emph{Journal of Geophysical Research: Solid Earth}, vol. 117, no.~B3, 2012.

\bibitem{wech2008automated}
A.~G. Wech and K.~C. Creager, ``Automated detection and location of cascadia tremor,'' \emph{Geophysical Research Letters}, vol.~35, no.~20, 2008.

\bibitem{schmidt2010source}
D.~Schmidt and H.~Gao, ``Source parameters and time-dependent slip distributions of slow slip events on the cascadia subduction zone from 1998 to 2008,'' \emph{Journal of Geophysical Research: Solid Earth}, vol. 115, no.~B4, 2010.

\bibitem{rogers2003episodic}
G.~Rogers and H.~Dragert, ``Episodic tremor and slip on the cascadia subduction zone: The chatter of silent slip,'' \emph{Science}, vol. 300, no. 5627, pp. 1942--1943, 2003.

\end{thebibliography}

 




\vfill

\end{document}